\documentclass[reprint,amsmath,amssymb,aps,prl]{revtex4-1}

\usepackage{graphicx}
\usepackage{caption}
\usepackage{subcaption}
\usepackage{hyperref}
\usepackage{float}

\floatstyle{ruled}
\newfloat{pseudo}{h}{lop}
\floatname{pseudo}{Algorithm}

\usepackage{color}

\newcommand{\p}{\mbox{\,.}}

\newcommand{\R}{\mathbb{R}}
\newcommand{\cN}{{\cal N}}

\newcommand{\tab}{\hspace*{5mm}}

\newcommand{\run}[1]{\includegraphics[width=0.15\linewidth]{./img/runs/{#1}.png}}

\newcommand{\bx}{\boldsymbol x}
\newcommand{\by}{\boldsymbol y}
\newcommand{\bX}{\boldsymbol X}
\newcommand{\bY}{\boldsymbol Y}
\newcommand{\bo}{\boldsymbol 1}

\newcommand{\bt}{\boldsymbol t}
\newcommand{\bA}{\boldsymbol A}
\newcommand{\bmu}{\boldsymbol \mu}
\newcommand{\bSig}{\boldsymbol \Sigma}
\newcommand{\bI}{\boldsymbol I}
\newcommand{\bR}{\boldsymbol R}
\newcommand{\bZ}{\boldsymbol Z}
\newcommand{\bz}{\boldsymbol z}

\newcommand{\bw}{\boldsymbol w}
\newcommand{\bv}{\boldsymbol v}
\newcommand{\bP}{\boldsymbol P}
\newcommand{\bT}{\boldsymbol T}
\newcommand{\bU}{\boldsymbol U}
\newcommand{\bS}{\boldsymbol S}
\newcommand{\bV}{\boldsymbol V}

\begin{document}

\preprint{APS/123-QED}

\title{An Expectation-Maximization Algorithm for the Fractal Inverse Problem}

\author{Peter Bloem}
\email{vu@peterbloem.nl}
\affiliation{Knowledge Representation and Reasoning Group\\
        VU University Amsterdam \\
		De Boelelaan 1105, 1081 HV Amsterdam, NL}

\author{Steven de Rooij}
\email{s.derooij@gmail.com}
\affiliation{Mathematical Institute \\
        University of Leiden\\
        Niels Bohrweg 1, 2333 CA Leiden, NL}

\date{\today}

\begin{abstract}
\noindent We present an Expectation-Maximization algorithm for the \emph{fractal inverse problem}: the problem of fitting a fractal model to data. In our setting the fractals are Iterated Function Systems (IFS), with \emph{similitudes} as the family of transformations. The data is a point cloud in ${\mathbb R}^H$ with arbitrary dimension $H$. Each IFS defines a probability distribution on ${\mathbb R}^H$, so that the fractal inverse problem can be cast as a problem of parameter estimation. We show that the algorithm reconstructs well-known fractals from data, with the model converging to high precision parameters. We also show the utility of the model as an approximation for datasources outside the IFS model class. 
\end{abstract}

\maketitle

\noindent Fractals are mathematical objects; commonly formalised as a class of sets or of probability distributions. What, in general, constitutes a fractal is not precisely defined \footnotemark, but certain properties are common to most examples; for instance self-similarity, infinitely fine structure, and a non-integer dimension.

\footnotetext{Mandelbrot originally defined fractals using Haussdorf dimension, but later stated that he preferred the word to be not precisely defined.}

\begin{figure}[b]
  \includegraphics[width=\linewidth]{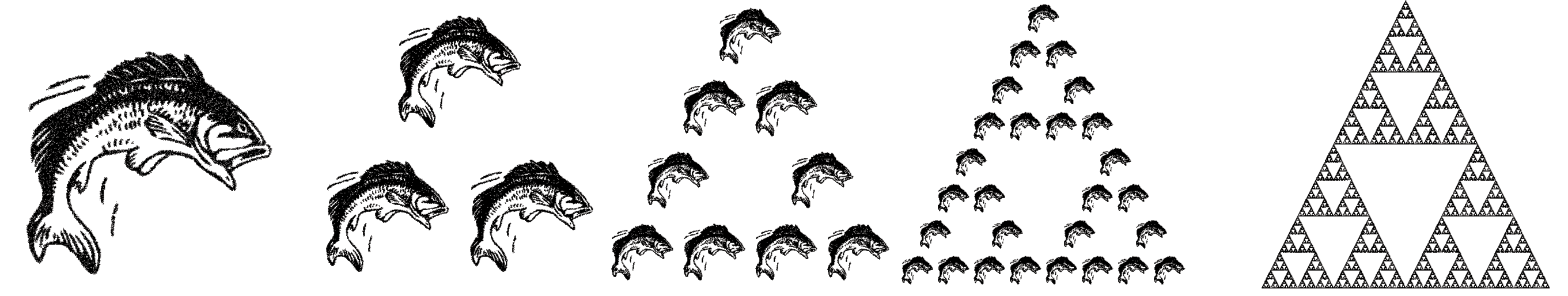}
  \caption{An example of an IFS.}
  \label{figure:example}
\end{figure}

Since they were named in the 1970s, fractals have increasingly been seen as a potential model for many natural phenomena. Mandelbrot put it as follows \cite{mandelbrot1982fractal}: 
\begin{quote}
\noindent Clouds are not spheres, mountains are not cones, coastlines are not circles, and bark is not smooth, nor does lightning travel in a straight line.
\end{quote}
Fractals have been used in many fields, including physics \cite{mandelbrot1984fractals}, geology \cite{cheng1997multifractal}, biology \cite{goldberger1992fractal} and economics \cite{turiel2003multifractal}.

One of the greatest problems with fractal analysis has always been the difficulty of finding a fractal model, given some observations. It may be visually clear that a cloud or a coastline `looks' fractal, and we may be able to determine that it has a non-integer dimension \cite{theiler1990estimating}, but how do we get from a dataset to a model? This is called the \emph{fractal inverse problem}.

Current approaches tend to rely on evolutionary algorithms \cite{deliu1991genetic,collet2000polar,nettleton1994evolutionary}. Such algorithms are expensive, and it can be difficult to get them to converge to precise models, even if the dataset itself is sampled from a fractal model. Other approaches are highly domain-specific, such as fractal image compression \cite{hart1996fractal}, which does not generalize well to data of arbitrary dimension. Some interesting results have been also been derived from the method of moments \cite{rinaldo1994inverse} and sampling random transformations from the data \cite{hart1997similarity}, but apart from the evolutionary methods, we are not aware of any other generic, practical methods. 

We focus on the family of \emph{Iterated Function Systems} (IFSs) \cite{barnsley2014fractals,hutchinson2000deterministic}, a broad class of fractals, capturing many well-known examples. Figure~\ref{figure:example} shows the basic principle: we start with some initial image and apply a small number, $K$, of contracting transformations to it, resulting in $K$ scaled down copies of the initial image. As we iterate this process, the image converges to a fractal. Which fractal emerges is entirely determined by the chosen transformations. By fixing a family of transformations $\cal F$, we define a family of fractals, each determined by a set of $K$ transformations chosen from $\cal F$: its \emph{components}. Here, we choose \emph{similitudes}, a subset of affine maps. Similitudes offer a good trade-off between expressiveness, and efficient optimization. 

The IFS concept generalizes naturally to \emph{probability distributions}: we choose the components of the IFS, and instead of applying them to an initial image, we apply them to an initial \emph{distribution}, combining these into a mixture of scaled-down copies of the original distribution. The iteration now results in a fractal probability distribution. This allows us to frame the fractal inverse problem as a problem of parameter estimation: for a set of points $\{x\} \subset \R^H$, find a set of $K$ similitudes, such that the likelihood of $\{x\}$ under the resulting IFS distribution is maximal. 

We present an Expectation-Maximization (EM) algorithm \cite{dempster1977maximum} to find this model. Since the specific sequence of components that ``generated'' a point $x$ in the dataset is unknown, we cast this information as a \emph{latent variable} and iterate between optimizing the latent variables given the model, and optimizing the model given the latent variables. 

We show that our algorithm reconstructs known fractals. We also apply the algorithm to some datasets sampled from images, and some natural data of higher dimension, showing that, while there are no IFSs to perfectly capture these images, the self similarity in the model does often allow a good fit, comparable to that of a mixture-of-Gaussians model. An open-source implementation is available \footnote{\url{https://github.com/pbloem/ifsem}}.

\paragraph{Notation} 

Let $\{\bx_i\}_{i\in[1,N]} \subset \R^H$ be our dataset. Let $\bX$ be the $N\times H$ matrix with $\bx_i$ as its columns. 

For $S \subseteq \R^H$, let $V(S)$ be a probability distribution on $\R^H$ with p.d.f. $v(\bx)$. Let $f_{\bt, \bA}(\bx) = \bA\bx + \bt$ be an invertible affine transformation. Then the transformation of $V$ by $f_{\bt, \bA}$ is defined by the relation $f_{\bt, \bA}(V)(S) = V({f_{\bt, \bA}}^{-1}(S))$. The density function of $f_{\bt, \bA}(V)$ is $f_{\bt, \bA}(V)(\bx) = |\bA^{-1}| v({f_{\bt, \bA}}^{-1}(\bx))$.

A \emph{Gaussian}, on $\R^H$ is determined by a mean $\bmu \in \R_H$ and a covariance matrix $\bSig \in \R^{H\times H}$, with $\cN_0 = (\boldsymbol 0, \bI^H)$. A \emph{spherical} or \emph{isometric} Gaussian has $\bSig = s\bI$ for some scalar $s$. Let $\bx$ be a random variable with $\bx \sim \cN(\bmu, \bSig)$, then $f_{\bt, \bA}(\bx) \sim \cN(\bt + \bA\bmu, \bA\bSig \bA^T)$.
A \emph{similitude} $f:\R^H \to \R^H$ is defined by a rotation matrix $\bR$, vector $\bt$ and scalar $s$ as follows $f_{\bt, \bR, s}(\bx) = s\bR \bx + \bt$.

Transforming a spherical Gaussian by a similitude yields
\begin{align}
&f_{\bt, \bR, s}\left(\cN(\bt_0, {s_0}^2 \bI)\right)(\bx) = \notag \\
&(2\pi)^{-\frac{H}{2}} ({ss_0})^{-H} \exp \left[- \frac{1}{2{s_0}^2s^2}\left\| \bx - (s \bR \bt_0 + \bt)\right \|^2\right] \label{line:inverse}
\end{align}

\paragraph{The IFS Model} 

\begin{figure}[bt]
  \center
  \includegraphics[width=\linewidth]{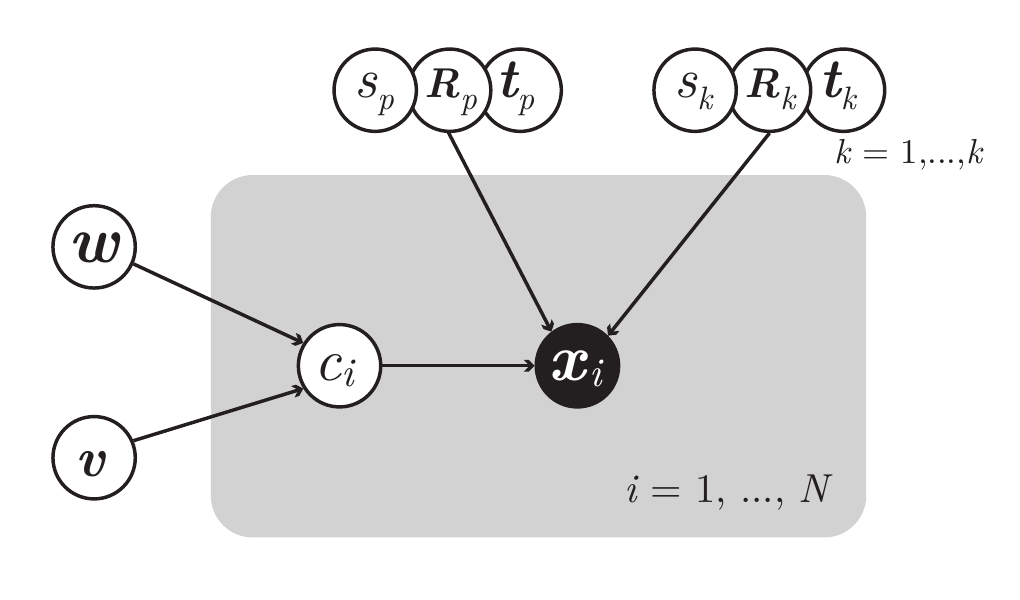}
  \caption{A graphical model, illustrating the components of the IFS model.}
  \label{figure:ifs-diagram}
\end{figure}

We define an \emph{Iterated Function System} of order $K$ and dimension $H$ as a pair $(\{f_k\}, \{w_k\})$ of $K$ similitude \emph{components} $f_k: \R^H \to \R^H$ with $K$ associated \emph{weights} $w_k$, nonnegative scalars, with $\sum_i w_i = 1$.  

Let $\bx_0 \sim p_0$, with $p_0$ any distribution on $\R^H$ with compact support. Define $\bx_{n+1}$ as the random variable with a mixture distribution over the $K$ random variables $f_k(\bx_n)$, with mixture weights $w_k$. We note two important properties of IFSs. First, the distribution on $\bx_D$ converges with $D$. We call the distribution it converges to the \emph{limit distribution}. Second, the limit distribution is independent of the choice of $p_0$. Thus, we can see the weights and components as \emph{parameters} of the limit distribution. For formal proofs and generalizations, see \cite{hutchinson2000deterministic}.

To make the model easier to fit to natural data, we extend the IFS model with some additional factors, as illustrated in Figure~\ref{figure:ifs-diagram}.
\begin{description}
\item[the components $\{(s_k, \bR_k, \bt_k)\}_k$] The $K$ similitudes that make up the model: $f_k(x) = s_k\bR_k\bx + \bt_k$.
\item[the weights $\bw$] A length-$K$ probability vector. 
\item[the depth weights $\bv$] A length-$D$ probability vector.
\item[the post-transform $s_p, \bR_p, \bt_p$] A single similitude.
\item[the code $c_i$] A sequence of integers $c_i = \langle c_{i1}, \ldots, c_{id} \rangle$.
\item[the data $\bx_i$] A single point in $\R^H$.
\end{description}
The first four form the \emph{parameters} $\theta$ of the model. The rest are \emph{observed} variables $\bx_i$ and the \emph{latent} variables $c_i$.  

These combine into a model as follows. Let $\{c_j\} = [1, K]^{[0, D]}$, the set of all sequences of integers from $[1, K]$ of length up to $D$, including the empty sequence. We define the function $f_j:\R^H \to \R^H$ as the composition of the post-transform and the components indicated by the code $c_j$:$f_j = f_p \circ f_{c_{j1}} \circ \ldots \circ f_{c_{jd}}$. We can now write the p.d.f. of the model as:
\begin{align*}
p(x\mid \theta) &= \sum_{j=1}^{|[1, K]^{[0, D]}|} \bv_{|c_j|} \left[\prod_i \bw_{c_{ji}}\right ] f_j(\cN_0)(x) \p
\end{align*}
That is, a mixture of $|[1, K]^{[0, D]}|$ Gaussians, where the $j$-th Gaussian has the weight $\bv_{|c_j|} \prod_i \bw_{c_{ji}}$. Equivalently: we evaluate the IFS to all depths from $0$ to $D$ and mix these with the weights in $\bv$.

Mixing different depths makes the model a generalization of good fall-back models: with $\bv_0 = 1$, the model becomes a spherical Gaussian. With $\bv_1 = 1$, the model becomes a mixture of K Gaussians \footnotemark[2]. Mixing depths also provides a ``gentle start.'' Most IFSs have a support with low dimension: i.e. almost all of $\R^H$ has probability zero. Thus, the slightest change in parameters means the difference between maximal and zero likelihood. The variable depth allows the search algorithm to start with low-fitness models with a smooth error surface, and slowly converge to the more challenging, deeper models.

The post-transform caters for data that is not properly centered. While we can make off-center IFSs by tuning the components, such models will not overlap properly at different depths. The post-transform allows us to learn a centered IFS in place, and map it to the data.

\footnotetext[2]{To generalize to generic MOG models, the components should be extended to positive definite affine functions.}

\begin{figure}[tb]
  \centering
  \begin{subfigure}[b]{0.47\linewidth}
    \includegraphics[width=\textwidth]{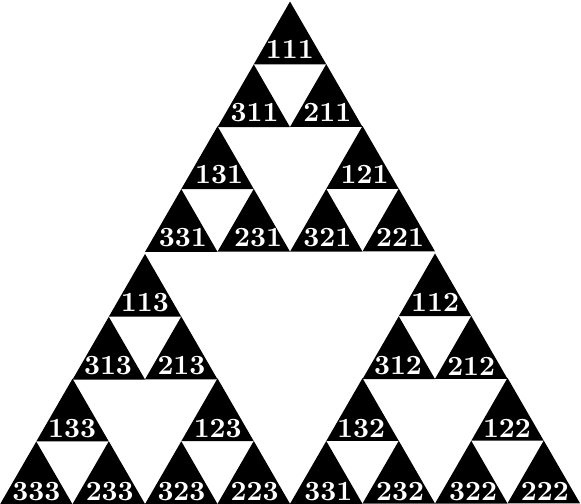}
    \caption{}
    \label{fig:sierpinski-codes}
  \end{subfigure}  
  \hspace{0.015\textwidth}
  \begin{subfigure}[b]{0.47\linewidth}
    \includegraphics[width=\textwidth]{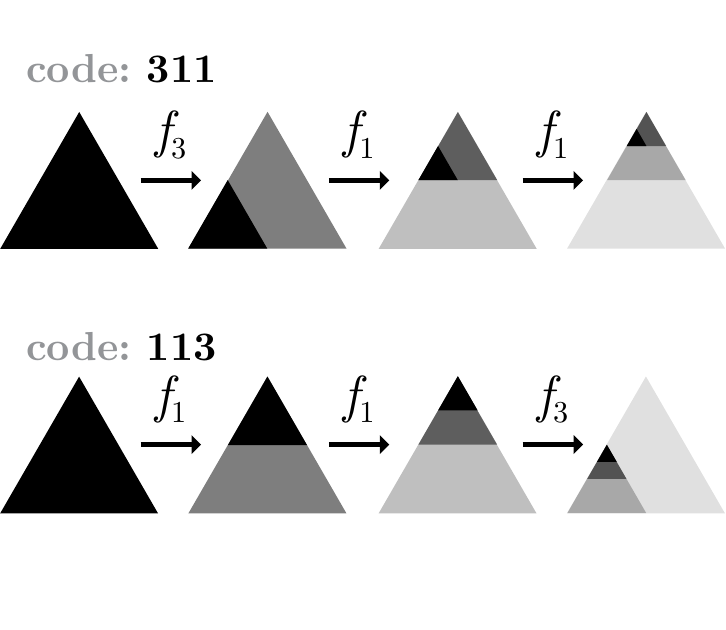}
    \caption{}
    \label{fig:code-construction}
  \end{subfigure}

  \caption{\small (a) Codes of length three on the Sierpinski triangle and the subsets they code for. (b) The construction of a subset from its code.}
  \label{figure:codes}
\end{figure}

\paragraph{The EM Algorithm}
\label{section:algorithm}

Our task is to select the parameters $\theta$ such that he probability of our data $\bX$ under the model $\theta$ is maximized. Unfortunately, we do not know the code $c_i$ corresponding to each point $\bx_i$. A common way to deal with such \emph{latent variables} is the EM algorithm. In brief, it starts with an initial model, computes the most likely values of the latent variables for each datapoint, and then re-computes the most likely model given these values of the latent variables, iterating until convergence.

The most concise way to describe the EM algorithm is in terms of a \emph{Q-function}:
\[
Q(\theta) = \sum_{\bZ} p(\bZ \mid \bX, \theta^\text{old}) \ln p(\bX, \bZ \mid \theta) \label{line:q-function}
\]
Where the sum iterates over all possible values of all latent variables $\bZ$. Note that the variable $\theta^\text{old}$, our existing estimate for the model parameters, is a constant. The \emph{expectation} step computes $p(\bZ \mid \bX, \theta^\text{old})$, and the \emph{maximization} step computes the $\theta$ that maximizes $Q(\theta)$. 

We first rewrite to:
\begin{align*}
Q(\theta) &= \sum_{i} p(\bz_i \mid \bx_i, \theta^\text{old}) \ln p(\bx_i \mid \bz_i, \theta) p(\bz_i \mid \theta) \\
     &= \sum_{i=1,j=1}^{N, M} \bP_{ij} \ln \cN_j(x) \;v_{|c_j|}\; \prod_{k\in c_j}w_k 
\end{align*}
with $M = |[1, K]^{[0, D]}|$ and $\bP$ an $N\times M$ matrix with 
\[
\bP_{ij} = p(c_j \mid x_i) = \frac{p(x_i\mid c_j) p(c_j) }{\sum_{a \in [1,K]^{[0, D]}} p(x_i\mid c_a) p(c_a)} \p 
\]
The computation of $\bP$ is the expectation step of the algorithm: $\bP_{ij}$ represents the \emph{responsibility} that the $j$-th Gaussian takes for the $i$-th datapoint. For the maximization step we optimize $Q$ with respect to the various elements of $\theta$, taking the partial derivative and setting it equal to zero. For clarity of notation, when optimizing for a certain subset $q$ of the parameters $\theta$ we write $Q(q)$, and omit any terms any multipliers constant in $q$.

TTo optimize $f_k$ and $f_p$, we must find $K$ maps and weights, and a post-transformation $f_p$, such that all the $M$ endpoint distributions provide maximal likelihood to their assigned points. The problem is that each term in $Q$ is a complicated mix of multiple components.

To make the optimization of $Q$ practical, we simplify the task in two ways. First, we optimize $f_p$ and $(\{f_k\}, \{w_k\})$ separately, taking the parameters not being optimized from $\theta^\text{old}$. Second, we simplify the $Q$-function. We first rewrite it as follows: Let $kc_j$ be the code $\langle k, c_{j1}, \ldots\rangle$ and let $M^- = |[1, K]^{[0, D-1}]|$. Let $\bY = {f_p^\text{old}}^{-1}(\bX)$. Then:
\begin{align*}
Q&(\{f_k\}, \{w_k\})  \\ 
&= \frac{1}{s_p}\sum_k \sum_{i, j} &\bP^k_{ij} \ln \left[ f_k(\cN_{j})(y_i) \; v_{|c_j|+1} p(c_j) \; w_k \right]
\end{align*}
with $\bP_{ij}^k = p(kc_j \mid \bx_i)$.
 
We have now written $Q$ in a form that isolates only the first component in the code. Of course, each $\cN_j$ in this sum still depends on all the components in the code $c_j$, but this is where we simplify the function: we take $\cN_j$ to be a \emph{constant}, computed from $\theta^\text{old}$, and optimize only for the first component in the code. This gives us, for the $k$-th component: $Q(f_k) = \sum_{i, j} \bP^k_{ij} \ln f_k(\cN_j)(y_i)$.

Using (\ref{line:inverse}), we rewrite $Q(f_k)$ as a mixture of transformations of $\cN_0$:
\begin{align*} 
Q(s_k, \bR_k, \bt_k) =& - p^k H \ln s_k - \\ 
&\sum_{i, j} \bP^k_{ij} \frac{1}{2{s_j}^2{s_k}^2} \left\|y_i-\bt_k - s_k \bR_k \bt_j \right\|^2 
\end{align*}
where $s_j$, $\bR_j$ and $\bt_j$ are the parameters of the similitude $f_j = f_p \circ f_{c_{j1}} \circ \ldots \circ f_{c_{jd}}$. 

To find $\hat\bt_k$, we solve $\partial Q(\bt_k)/\partial \bt_k = 0$:
\begin{align*}
\hat \bt_k &= \frac{1}{p^k_z}\bY\bP^k\bZ\bo - \frac{1}{p^k_z}s_kR_k\bT\bZ{\bP^k}^T\bo = \by^k - s_k\bR_k \bt^k
\end{align*} 
where $\bT$ is the matrix with $\bt_j$ as its columns, $\bZ = \text{diag}({s_1}^{-2}, \ldots, {s_M}^{-2})$ and ${p^k_z} = \bo^TP^kZ\bo$.

Finding the optimal rotation matrix $\hat\bR_k$ is complex, since it must be a \emph{rotation} matrix. We first rewrite the objective function to the form $\text{tr}(\bA^T\bR_k)$, for some $\bA$:
\[ 
Q(\bR_k) = \text{tr}\left(\left[\bY^k\bP^k\bZ{\bT^k}^T\right]^T\bR_k\right) \text{,}\;\;
\begin{aligned}
\bY^k &= \bY - \by^k {1}^T\\
\bT^k &= \bT - \bt^k {1}^T
\end{aligned}
\]
The optimal $\bR_k$ can then be derived from the singular value decomposition (SVD) of $\bA=\bY^k\bP^k\bZ{\bT^k}^T$: if $\bA = \bU\bS\bV^T$, with $\bU$, $\bS$ and $\bV$ defined as normal for the SVD then we have $\hat \bR_k = \bU \;\text{diag} (1, \ldots, 1, |\bU\bV^T|)\; \bV^T$ \cite{myronenko2009closed}.

Finally, we derive the scaling parameter $s_k$ by filling in $\hat \bt_k$ and solving $\partial Q(s_k)/\partial s_k = 0$.

The optimization of the post-transform follows the same pattern. The optimal values for the weights $\bw$ and the depths $\bv$ follow from straightforward differentiation under constraints. The complete update rules are provided in Figure~\ref{figure:algorithm}. For detailed derivations, see \cite{bloem2016single}. The derivations of $\hat\bt_k$ and $\hat\bR_k$ are inspired by \cite{myronenko2010point}.

\begin{figure}[bt]
{
\hrule
\vspace{3mm}
\raggedright
% Given: a dataset $\bX$, a number of components $K$, a maximum depth $D$. \\
%\tab \\
$\bP_{ij} \leftarrow \cN_j(\bx_i) \bv_{|c_j|} \prod_{a \in c_j} \bw_a$ \hfill \emph{\# Expectation step}\\ 
Normalize $\bP$ so that $\bP\bo = \bo$ \\
\textbf{for each} $k \in [1,K]$: \\
\tab $\bP^k \leftarrow$ submatrix of $\bP$'s columns $j$ for which $c_{j1} = k$\\
\tab \\
$\bY \leftarrow {f^\text{old}_p}^{-1}(\bX)$ \hfill \emph{\# Maximization step}\\
for all $d$, $\hat\bv_d \propto \sum_{j : |c_j| = d} \bP_{ij} $\\
\textbf{for each} $k \in [1,K]$: \\
\tab $\hat\bw_k \leftarrow \bo^T \bP^k \bo / \sum_i {\bo^T \bP^i \bo}$\\
\tab $\by^k \leftarrow {p^k}^{-1}\bY\bP^k\bZ\bo, \,\,\, \bt^k \leftarrow {p^k}^{-1} \bT \bZ{\bP^k}^T \bo$\\
\tab $\bY^k \leftarrow \bY + \by^k\bo^T$, \,\,\, $\bT^k \leftarrow \bT^k + \bt^k\bo^T$\\
\tab $\bZ \leftarrow \text{diag}({s_1}^{-2}, \ldots, {s_{M^-}}^{-2})$\\
\tab $\bU,\bS,\bV^T \leftarrow \text{svd}(\bY^k\bP^k\bZ{\bT^k}^T)$\\ 
\tab $\hat\bR_k \leftarrow \bU\;\text{diag}(1,\ldots,1,|\bU\bV^T|) \bV^T$ \\
\tab $\hat s_k$: solve ${s_k}^{-2}\; \text{tr}({\bY^k}^T\text{d}(\bP^k\bZ\bo)\bY^k)$ \\
\tab \tab $+ {s_k}^{-1}\;\text{tr}(\bT^k \bZ{\bP^k}^T{\bY^k}^T\bR_k) - Hp^k= 0$ \\
\tab $\hat\bt_k \leftarrow \by^k - s_k \bR_k \bt^k$ \\
\tab \\
$\bX^p \leftarrow X - \bx^p\bo^T$, $\bT^p \leftarrow \bT + \bt^p\bo^T$\\
$\bZ \leftarrow \text{diag}({s_1}^{-2}, \ldots, {s_M}^{-2})$\\
$\bU,\bS,\bV^T \leftarrow \text{svd}(\bX^p\bP\bZ{\bT^p}^T)$\\ 
$\hat\bR_p \leftarrow \bU\;\text{diag}(1,\ldots,1,|UV^T|) V^T$ \\
$\hat s_p$: solve \\
\tab ${s_p}^{-2}\text{tr}(\bX^T\text{d}(\bP\bZ\bo)\bX) + {s_p}^{-1} \text{tr}(\bT \bZ\bP^T\bX^T\bR_p) - H p = 0$ \\
$\hat\bt_p \leftarrow \bx^p - s_p\bR_p\bt^p$ \\
\vspace{3mm}
\hrule
}
\caption{One iteration of the IFS-EM algorithm.}
\label{figure:algorithm}
\end{figure}

\paragraph{Results}
\label{section:results}

\begin{figure*}[tbh]
  \centering
   \begin{subfigure}{0.5\linewidth}
	\begin{tabular}{c c c c c c}
		 data  & it 1 & it 20 & it 300 & full \\
		 \run{sierpinski/data}  & \run{sierpinski/best.000000} & \run{sierpinski/best.000019} &  \run{sierpinski/best.000299} & \run{sierpinski/best.000299.deep} \\
		 \run{sierpinski-off/data}  & \run{sierpinski-off/best.000000} & \run{sierpinski-off/best.000019} & \run{sierpinski-off/best.000299} & \run{sierpinski-off/best.000299.deep} \\
		\run{koch2/data} & \run{koch2/best.000000} & \run{koch2/best.000019} & \run{koch2/best.000299} & \run{koch2/best.000299.deep} \\
		 \run{koch4/data}  & \run{koch4/best.000000} & \run{koch4/best.000019} & \run{koch4/best.000299} & \run{koch4/best.000299.deep} \\
		 \run{square/data}  & \run{square/best.000000} & \run{square/best.000019} &  \run{square/best.000299} & \run{square/best.000299.deep}  \\
		 \run{coast/data}& \run{coast/best.000000} & \run{coast/best.000019} & \run{coast/best.000299} & \run{coast/best.000299.deep}  \\
		 \run{broccoli/data} & \run{broccoli/best.000000} & \run{broccoli/best.000019} & \run{broccoli/best.000299} & \run{broccoli/best.000299.deep}  \\
		 \run{sphere/data} & \run{sphere/best.000000} & \run{sphere/best.000019} &  \run{sphere/best.000299} & \run{sphere/best.000299.deep} \\
	\end{tabular}   
	\caption{}
	\label{figure:twod}
  \end{subfigure}  
  \begin{subfigure}{0.49\linewidth}
  	\begin{subfigure}{\linewidth}  
	  	\begin{tabular}{ r r r r r r r}
	  		data & $c$ & 0 & 10 & 50 & 100 & 500 \\
			\hline
			sierpinski & 3 & 0.05 & 0.10 & 0.21 & 0.21 &  0.63 \\
			sierp. (nu) & 3 & 0.04 & 0.68 & 0.95 & 0.99 & 1.00 \\
			koch  & 2 & 0.16 & 0.27 & 0.30 & 0.33 & 0.48 \\
			koch  & 4 & 0.26 & 0.72 & 0.54 & 0.37 & 0.35 \\
			square & 4 & 0 & 0.32 & 0.42 & 0.60 & 0.61 \\
			\hline
		\end{tabular}
		\caption{}
		\label{figure:convergence}
	\end{subfigure}
	\begin{subfigure}{\linewidth}
		\begin{subfigure}{0.31\linewidth}  
			\includegraphics[width=\linewidth]{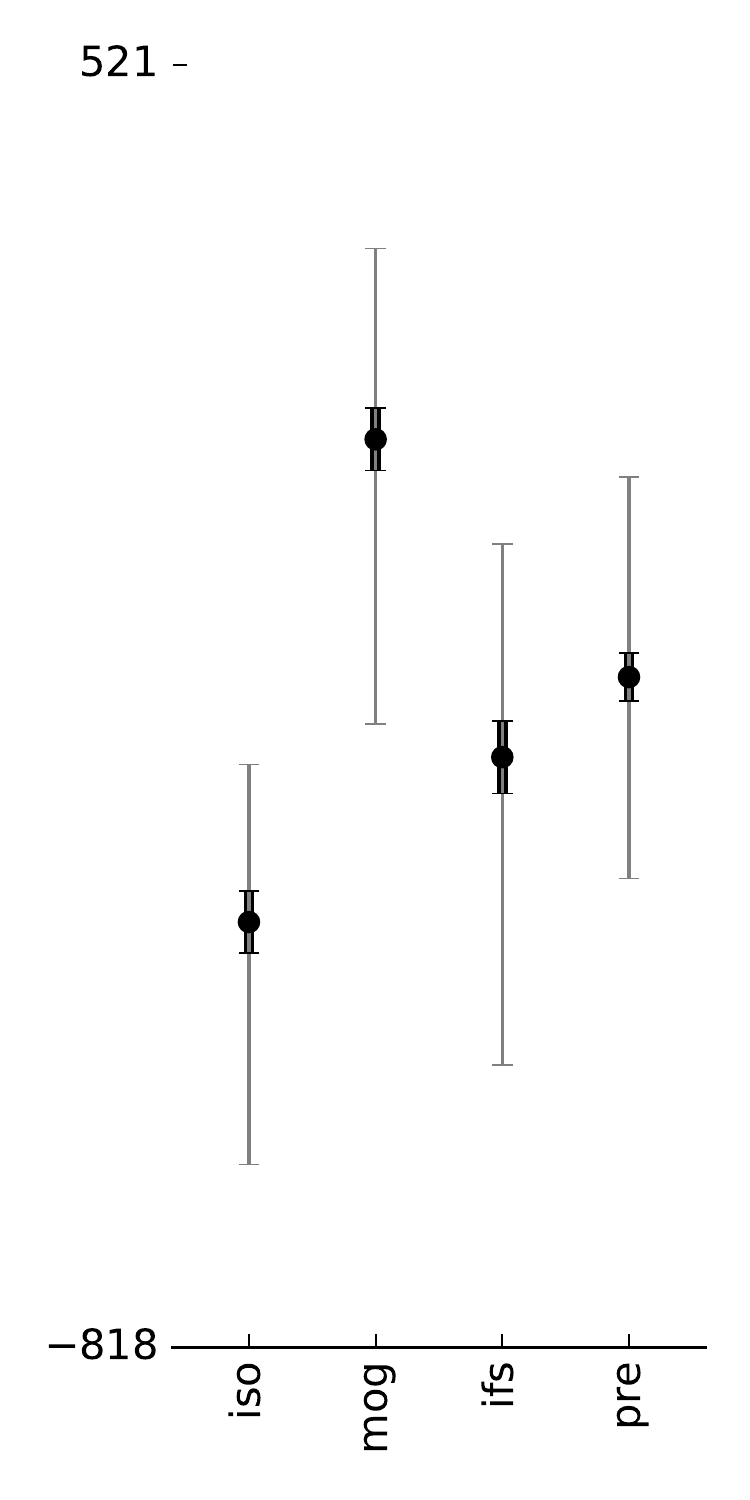}\\
			currency
		\end{subfigure} 
		\begin{subfigure}{0.31\linewidth}  
			\includegraphics[width=\linewidth]{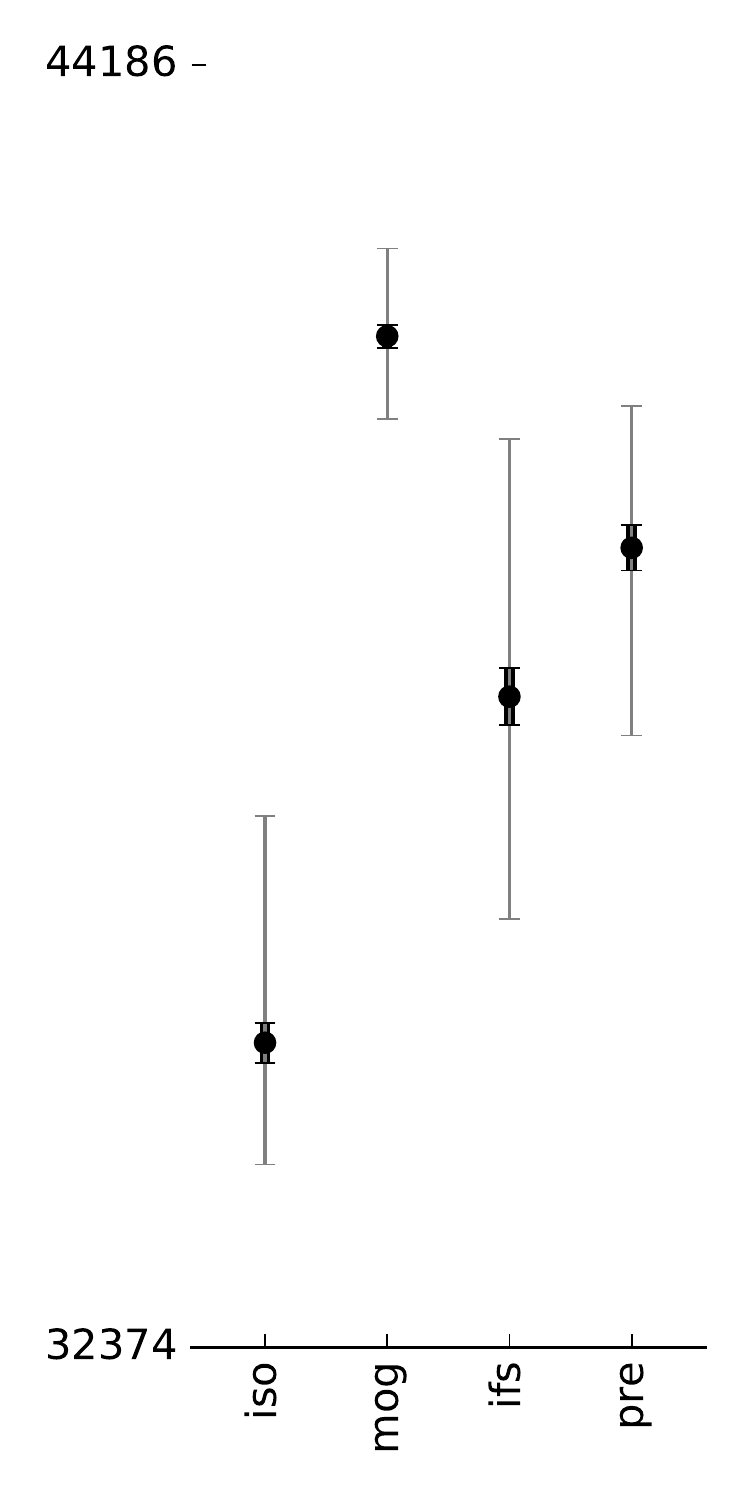}\\
			gait
		\end{subfigure} 		
		\begin{subfigure}{0.31\linewidth}  
			\includegraphics[width=\linewidth]{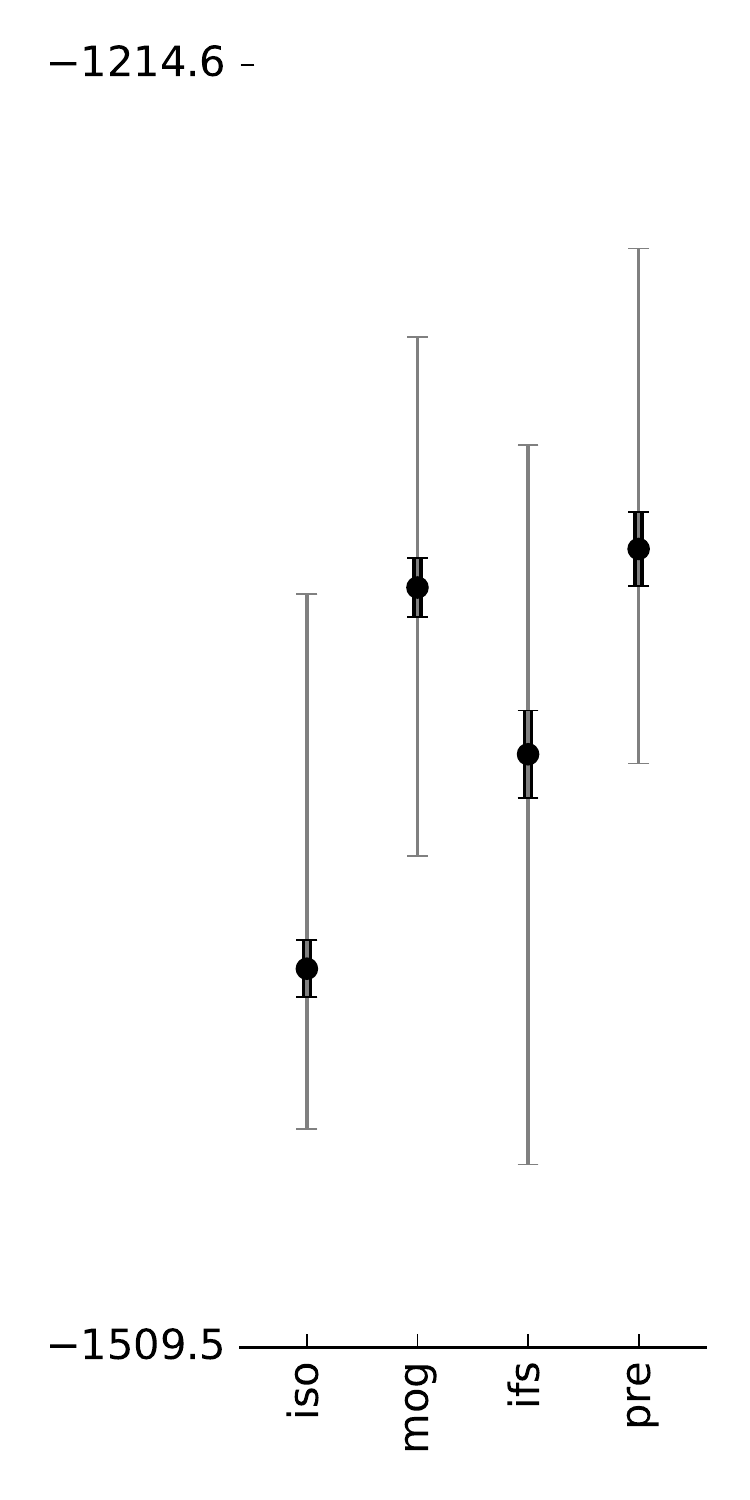}\\
			galaxies
		\end{subfigure} 
		\caption{}		
	\end{subfigure} 
	\begin{subfigure}{\linewidth}
			\begin{subfigure}{0.31\linewidth}  
				\includegraphics[width=\linewidth]{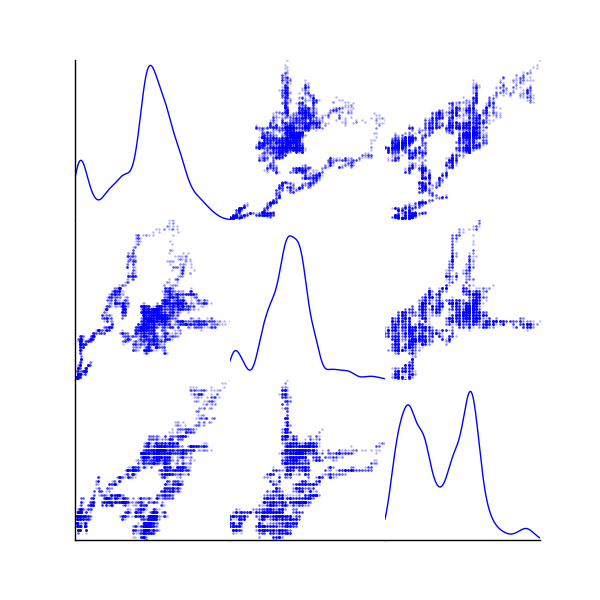}\\
				data
			\end{subfigure} 
			\begin{subfigure}{0.31\linewidth} 
				\includegraphics[width=\linewidth]{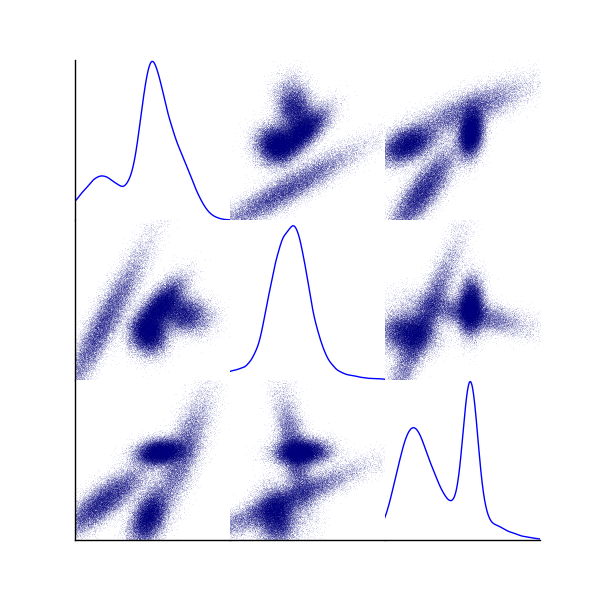}\\
				mog
			\end{subfigure}
			\begin{subfigure}{0.31\linewidth}  
				\includegraphics[width=\linewidth]{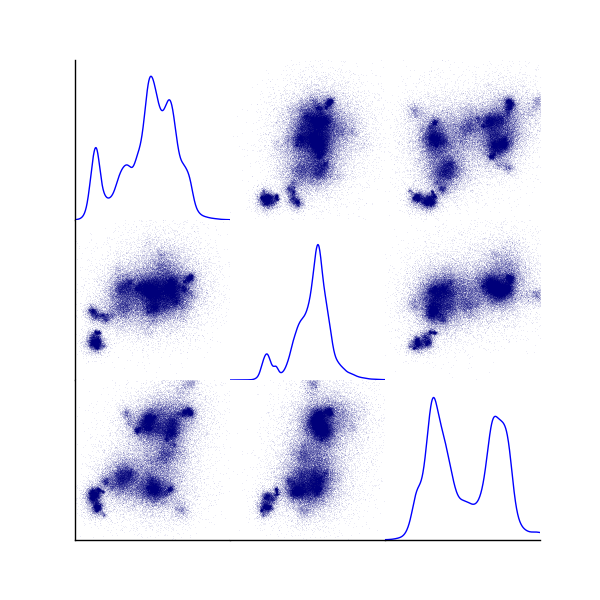}\\
				pre
			\end{subfigure}
			\caption{} 
			\label{figure:scatter}
	\end{subfigure} 
  \end{subfigure}
  \caption{\small (a) Runs on several 2D datasets: the Sierpinski triangle with uniform and non-uniform weights, the Koch curve with 2 and 4 components, a uniform distribution over a square, a section of coast line, shadows on a head of romanesco broccoli and a circle. The red frame represents the bi-unit square before the post transform is applied. Each component maps the red frame onto one of the blue frames. The last column shows a sample from the final model at infinite depth. (b) The proportion of 128 runs that convergence for a given dataset, for different numbers of initial candidates. (c) Likelihoods of the final model on three 3-dimensional datasets compared against two mixture-of-Gaussian models, one constrained to spherical Gaussians (iso), one with unconstraind Gaussians (mog). We report results for a run with a simple initial model (ifs), and one with candidate pre-selection (pre). Inner error bars show the standard error, outer error bars show the range. (3) Scatter matrices for the currency experiment.}
  \label{figure:results}
\end{figure*}

When fitting to data from an IFS model, it can be difficult to establish whether the algorithm has converged to the target. The same model can be built in many equivalent ways, and the likelihood under the true model grows unbounded with the depth. As a proxy for convergence, we use the model depth $\bv$. If at least 95\% of the weight is assigned to the deepest model, we consider the model to have converged. Deep models are highly unstable, so only an accurate model will converge to higher depths. 

For performance reasons, we subsample a minibatch of size $N'$ for each iteration.

The model is very sensitive to the choice of initial parameters. We use two methods to choose the initial model. The first generates $K$ points from a uniform distribution on the bi-unit $H$-sphere, and then constructs the $K$ components with these as fixed points. Each component has $s_k = \frac{1}{2}$ and $\bR_k$ chosen from the uniform distribution over rotation matrices. With this strategy we see visually accurate models emerging, with varying probability \cite{bloem2016single}. The convergence rate, however is low. The second initialization strategy, called \emph{pre-selection} remedies this. We use a pool candidate models, initialized as described above, each trained at depth 3 for 100 iterations with $N' = 500$. We select the model with the highest mean depth ($\sum_i iv_i$) as the initial model. 

First, we run the model on several 2D datasets (Figure~\ref{figure:twod}). In all cases, we use pre-selection with 10 candidates. We then search for 300 iterations, with $N' = 500$ and $D=6$. For each dataset, we show the run that resulted in the model with the highest likelihood (on withheld data), out of 8 repeats. For all experiments we set $K=3$, except for the koch curve ($K=2$ and $K=4$) and the square ($K=4$).

The second experiment (Figure~\ref{figure:convergence}) shows the influence of the candidate pool. We repeat the experiments described above 128 times with different sized pools. We report the proportion of runs that converged. In general, larger pools lead to higher convergence, with the exception of the 4-component Koch curve. This may be because mean-depth is not a good pre-selection criterion in this case.

In our final experiment, we apply the model to 3D data and compare it to basic MOG models. Specifically, we test a mixture of isometric Gaussians and an unconstrained MOG. We report the result of a basic IFS and a model with pre-selection on 100 candidates. We repeat each experiment 20 times.

The datasets are: a timeseries of currency conversion rates for three currencies (\cite{franses2000non,hyndman2010time} 4773 points), human motion data from an accelerometer worn around the trunk  (\cite{bachlin2010wearable}\footnote{We select only class-0 motions (i.e. natural motion, recorded in between the experiments).} 777052 points) and the large scale distribution of galaxy clusters in the observable universe (\cite{skrutskie2006two}\footnote{Preprocessed data downloaded from \url{https://www.cfa.harvard.edu/~dfabricant/huchra/seminar/lsc/}.} 5641 points). For the MOG models, the whole dataset was used each iteration. For the IFS models we used $N' = 10000$. In all cases, we trained for 100 iterations, with $K=4$ and $D=5$.

The IFS model outperforms the isometric MOG (which has similar parameter complexity). The unconstrained MOG model, which has more degrees of freedom than the IFS models, tends to outperform the IFS model. Clearly, the IFS model will not soon replace the MOG model as a general-purpose probability density model. However, there may be a niche of domains for which the IFS model is suitable. Figure~\ref{figure:scatter} shows the difference in strategies between the two models.

\paragraph{Discussion}
\label{section:discussion}

To our knowledge, this is the first solution to the fractal inverse problem that does not use a general-purpose optimization technique like genetic algorithms. 

The model easily generalizes to function classes more general than similitudes, although convex optimization or gradient descent may then be required for the maximization step. Future work includes developing a variational algorithm \cite{beal2003variational} to increase stability, and the extension of the algorithm to random IFSs \cite{hart1996fractal} to model non-deterministic fractals.

\paragraph{Acknowledgements} We thank Pieter Adriaans for valuable discussions in the early stages of this project. This publication was supported by the Dutch national program COMMIT, by the Netherlands eScience center, and by the Amsterdam Academic Alliance Data Science (AAA-DS) Program Award to the UvA and VU Universities.

\bibliography{fractal}

\end{document}